\documentclass[conference]{IEEEtran}
\ifCLASSINFOpdf
\else
\fi
\usepackage{amssymb}
\usepackage{amsmath,bm}
\usepackage{algorithm}
\usepackage{algorithmic}
\newcounter{DefNum}
\usecounter{DefNum}
\newcounter{ExaNum}
\usecounter{ExaNum}
\newcommand{\Def}{\noindent\textbf{Definition~\arabic{DefNum}.~}\refstepcounter{DefNum}}
\newcommand{\Exa}{\noindent\textbf{Example~\arabic{ExaNum}.~}\refstepcounter{ExaNum}}
\newcommand{\nmi}{\mathrm{NMI}}
\usepackage{graphicx}
\graphicspath{{./fig/}}
\usepackage{slashbox,pict2e}
\begin{document}

\title{Evidential Label Propagation Algorithm for Graphs}
\author{\IEEEauthorblockN{Kuang Zhou$^\text{a,b}$,
Arnaud Martin$^\text{b}$, 
Quan Pan$^\text{a}$, and
Zhun-ga Liu$^\text{a}$
}
\IEEEauthorblockA{a. School of Automation, Northwestern Polytechnical University,
Xi'an, Shaanxi 710072, PR China. }
\IEEEauthorblockA{b. DRUID, IRISA, University of Rennes 1, Rue E. Branly, 22300 Lannion, France}
}

\maketitle

\begin{abstract}
Community detection   has attracted considerable attention crossing many areas as it
can be used for discovering the structure and features of
complex networks. With the increasing size of social networks in real world,  community detection approaches should be  fast and accurate.
The Label Propagation Algorithm (LPA) is known to be one of the  near-linear solutions and benefits of easy implementation,
thus it forms a good basis for efficient community detection
methods.  In this paper, we extend the  update rule and
propagation criterion of LPA in the framework of belief functions. A new community detection approach, called Evidential Label
Propagation (ELP), is proposed as an enhanced version of conventional LPA. The node influence is first defined to guide
the propagation process. The plausibility is  used to determine the domain label of each node. The update
order of nodes is discussed to improve the robustness of the method. ELP algorithm will converge after
the domain labels of all the nodes become unchanged. The
mass assignments  are calculated finally as  memberships of nodes. The overlapping nodes and outliers
can be detected simultaneously through the proposed method. The experimental results  demonstrate
the effectiveness of ELP.
\end{abstract}

\begin{IEEEkeywords}
Label propagation, theory of belief functions, outliers, community detection.
\end{IEEEkeywords}

\IEEEpeerreviewmaketitle

\section{Introduction}
With the development of computer and Internet technologies, networks are everywhere
in our common life. Graph models are useful in describing and analyzing many different
kinds of relationships and interdependencies. In order to have a better understanding of
organizations and functions in real-world networked systems, community structure
of graphs  is a primary feature that should be taken into consideration. Communities, also called clusters or modules,
are groups of nodes (vertices) which probably
share common properties and/or play similar roles within the graph. They can extract
specific structures from complex networks, and consequently community detection has
attracted considerable attention crossing many areas from physics, biology, and economics to sociology, where
systems are often represented as graphs.

Recently, significant progress has been achieved in this research field and several popular algorithms for community detection have been  devised.
One of the most popular type of classical methods partitions networks by optimizing some criteria such as
the modularity measure (usually denoted by $Q$) \cite{newman2004finding}.
But recent researches have found that the modularity based algorithms could not detect communities smaller than a certain size.
This problem is famously known as the resolution limit \cite{fortunato2007resolution}.
Another family of approaches considers hierarchical clustering techniques. It merges or splits clusters according to a topological measure of similarity between the nodes and tries to  build a hierarchical tree of partitions \cite{kim2015detecting}. Some
other popular community detection approaches using spectral
clustering \cite{newman2013spectral} or partitional clustering methods \cite{zhou2015similarity} can be found.

As the size of analyzed networks grows rapidly, the complexity of community detection algorithms needs to be kept
close to linear. The Label Propagation Algorithm (LPA),  which was first investigated in \cite{raghavan2007near},  has the
benefits of nearly-linear running time and easy implementation,
thus it forms a good basis for efficient community detection
methods. It only uses the
network structure  and requires neither optimization of a predefined objective function nor prior information about the communities.
In this model every node is initialized with a unique label. Afterwards each node adopts the label that most of its neighbors currently
have at every step. In this iterative process densely connected groups of nodes form a consensus on a unique label to form communities.

The behavior of LPA is not stable  because of the randomness. Different communities may be detected in
different runs over the same network. Moreover,  by assuming
that a node always adopts the label of the majority of its
neighbors, LPA ignores any other structural information existing in the neighborhood of this
node. Another drawback for LPA is that it can only handle  disjoint and
non-overlapping communities.  However, in real  cases,
one member in a network might span multiple communities. For instance, one may naturally belong
to several social groups like friends, families, and schoolmates.

Although most of the nodes in a graph follow a common community distribution pattern, some certain objects may deviate significantly
from the pattern. It is of great value to detect such outliers in
networks for de-noising data thereby improving the quality of the detected community structure and also for
further analysis. Finding community outliers is an important problem
but has not received enough attention in the field of social network analysis.

The theory of belief functions, also called Dempster--Shafer Theory (DST), offers a mathematical framework for modeling uncertainty and imprecise
information \cite{ds2}.
It has been widely employed in various fields,
such as data classification \cite{denoeux1995k,liu2015new}, 
data clustering \cite{masson2008ecm,zhou2015evidential, zhou2016ecmdd},
social network analysis \cite{wei2013identifying,zhou2014evidential,zhou2015median}
and statistical estimation \cite{denoeux2013maximum,zhou2014evidentialem}. Belief functions are defined on the
power set of the frame which greatly enriches the expression power. The compound sets of the frame can be used
to describe the uncertain information and our ignorance.

In this paper, we enhance the original LPA by introducing  new update and propagation strategies.
A novel Evidential Label Propagation (ELP) algorithm is presented  to detect communities. The main contribution of this work can be summarized as:
\begin{itemize}
  \item The influence of each node to a target is defined considering both the similarities and local densities. The larger the influence of one
  node to the target node is,  the easier its label can be propagated to the target.
\item Based on the node influence, a new label propagation algorithm, named ELP, is proposed for graphs. The method for determining
  the update order of nodes is devised to improve the robustness of ELP.
  \item The Basic Belief Assignments (bbas) of nodes are defined for each detected communities in the framework of belief functions.
  The overlapping nodes  and  outliers can be detected simultaneously through the obtained bbas.
\end{itemize}

The remainder of this paper is organized as follows. In Section \ref{background}, some
basic knowledge is briefly introduced. The ELP algorithm is presented in detail in Section \ref{ELPpresent}.
In order to show the effectiveness of the proposed
community detection approach, in Section \ref{secexp} we test the ELP algorithm
 on different graph data sets and make comparisons with related methods.
Conclusions are drawn in the final section.

%
%
%
%
%
%
%

\section{Background}
\label{background}
In this section some related preliminary  knowledge will be presented. Some basis of belief function theory will
be recalled first, then two existing algorithms related to the proposed method will be briefly described.
\subsection{Theory of belief functions}
Let $\Omega=\{\omega_{1},\omega_{2},\ldots,\omega_{c}\}$ be the finite domain of
$X$, called the discernment frame. The belief functions are defined on the power
set $2^{\Omega}=\{A:A\subseteq\Omega\}$.

The function $m:2^{\Omega}\rightarrow[0,1]$ is said to be the Basic Belief
Assignment (bba) on $2^{\Omega}$, if it satisfies:
\begin{equation}
\sum_{A\subseteq\Omega}m(A)=1.
\end{equation}
Every $A\in2^{\Omega}$ such that $m(A)>0$ is called a focal element.
The credibility and plausibility functions are defined  in Eqs.$~\eqref{bel}$ and $\eqref{pl}$ respectively:
\begin{equation}
Bel\text{(}A\text{)}=\sum_{B\subseteq A, B \neq \emptyset} m\text{(}B\text{)} ~~\forall A\subseteq\Omega,
\label{bel}
\end{equation}

\begin{equation}
 Pl\text{(}A\text{)}=\sum_{B\cap A\neq\emptyset}m\text{(}B\text{)},~~\forall A\subseteq\Omega.
 \label{pl}
\end{equation}
Each quantity $Bel(A)$  measures the total support given to $A$, while $Pl(A)$ represents potential amount of support to $A$.
The two functions  are linked by the following relation:
\begin{equation}
  Pl(A) = 1 -Bel(\overline{A}), ~~\forall A \subseteq \Omega,
\end{equation}
where $\overline{A}$ denotes the complementary set of $A$ in $\Omega$. The function $pl: \Omega \rightarrow [0,1]$ that
maps each element $\omega_i$ in $\Omega$ to its plausibility $pl(\omega_i)=Pl(\{\omega_i\})$ is called the
contour function associated to $m$.

A belief function on the credal level can be transformed into a probability function by Smets method \cite{smets2005decision}, where the
mass $m(A)$ is
equally distributed among the elements of $A$.
This leads to the concept of pignistic probability, $BetP$, defined by
\begin{equation}
 \label{pig}
	BetP(\omega_i)=\sum_{\omega_i  \in A \subseteq \Omega } \frac{m(A)}{|A|(1-m(\emptyset))},
\end{equation}
where $|A|$ is the number of elements of $\Omega$ in $A$.

How to combine efficiently several bbas coming from distinct sources  is a major information fusion problem in the belief function
framework. Many rules have been proposed for such a task. When the information sources are reliable, several distinct
bodies of evidence characterized by different
bbas can be combined using Dempster-Shafer (DS) rule  \cite{ds2}. If bbas $m_j, j=1,2,\cdots,S$ describing $S$
distinct items of evidence on $\Omega$, the DS
rule of combination of $S$ bbas can be mathematically defined as
\begin{align}\label{dsrule}
   (m_1\oplus m_2 &\oplus\cdots \oplus m_S)(X)=  \nonumber \\ &\begin{cases}
    0 & \text{if}~ X = \emptyset,\\
     \frac{\sum\limits_{Y_1 \cap \cdots \cap Y_S = X} \prod_{j=1}^{S}m_j(Y_j)}{1-\sum\limits_{Y_1 \cap \cdots \cap Y_S = X} \prod_{j=1}^{S}m_j(Y_j)} & \text{otherwise}.
  \end{cases}
\end{align}
\subsection{E$K$-NNclus clustering}
Recently, a new decision-directed clustering algorithm for relational data sets is put forward based on
the evidential $K$ nearest-neighbor (E$K$-NN) rule~\cite{denoeux2015ek}.
Starting from an initial partition, the algorithm, called E$K$-NNclus, iteratively reassigns objects to clusters using
the E$K$-NN rule \cite{denoeux1995k}, until a stable partition is obtained. After convergence, the cluster membership of each
object is described by a Dempster–-Shafer mass function assigning a mass to each specific cluster and to the whole
set of clusters.
\subsection{Label propagation}
Let $G(V,E)$ be an undirected network, $V$ is the set of  $N$ nodes, $E$ is the set of edges. Each node $v(v \in V)$ has a label $c_v$.
Denote by $N_v$ the set of neighbors of node $v$.
The Label Propagation Algorithm (LPA) uses the network structure alone to guide its
process. It starts from  an initial configuration where every node has a unique label. Then at every step one node (in
asynchronous version) or each node (in a synchronous version) updates
its current label to the label shared by the maximum number of its neighbors. For node $v$, its new label
can be updated to $\omega_j$ with
\begin{equation}
  j = \arg \max_{l} \{|u:c_u=l,u \in N_v| \},
\end{equation}
where  $|X|$ is the cardinality of set $X$, and $N_v$ is the set of node $v$'s neighbors. When there are multiple
maximal labels among the neighbors’ labels, the new label is
picked randomly from them.  By this iterative process
densely connected groups of nodes form consensus on one label to form communities, and each node has more neighbors in its own
community than in any of other community.    Communities
are identified as a group of nodes sharing the same label.
\section{Approach}\label{ELPpresent}
Inspired from LPA and E$K$-NNclus, we propose here the ELP algorithm for community detection.
After an introduction of  the concept of node influence,  the whole ELP algorithm will be presented in detail.
Consider the network $G(V,E)$. Let the degree of node $i$ be $d_i$, and    $\bm{A}=(a_{ij})_{N\times N}$ denote
the adjacency matrix, where $a_{ij}=1$ indicates that there is a direct edge between nodes $i$ and $j$.
\subsection{The influence of nodes}
\label{nodeinf}

\Def The local density of node $i$ in graph $G$ can  be defined as
\begin{equation}
  \rho_i = \frac{d_i}{N-1}, i = 1,2, \cdots,N,
\end{equation}
where $N = |V|$ is the number of nodes in the graph.

The value of $\rho_i$ describes the importance of node $i$ to some extent. The nodes that are playing  central roles in the network
have relatively large local densities.

\Def Denote the influence  of the node $v$ to its neighbor
node $u$ by $\delta_{uv}$.
\begin{equation}\label{siminf}
    \delta_{uv} = sim(u,v) \left(\frac{\rho_v}{\rho_u}\right)^\eta,
\end{equation}
where  $sim(u, v)$ denotes the similarity between nodes $u$ and $v$. Parameter $\eta$ is
adjustable  and it can be set to 1 by default. Many similarity measures can be adopted  here. In this
paper, the simple Jaccard Index is adopted:
   \begin{equation}\label{Jaccard}
            sim(u,v)=\frac{|N_u \cap N_v|}{|N_u \cup N_v|},
        \end{equation}
where $N_x=\{w\in V\setminus x: a(w,x)=1\}$ denotes the set of vertices that are adjacent to node $x$.

It should be noticed that the value of $\delta_{uv}$ is not equal to $\delta_{vu}$. In fact, we want to model the label propagation process
according to $\delta_{uv}$. The larger  the influence of node $v$ to node $u$ is,
the larger possibility that node $u$ will adopt the label of node $v$.
It is similar to the information propagation on social networks, where we are more likely to believe an  authority who is usually a center or an
important member in the community.
\subsection{Evidential label propagation}
In the original LPA, when updating the label of node $i$, the number of neighbors belonging to each class is counted, and  the label with maximal
frequency is adopted. In this case, the importance of each node in the neighborhood is considered equal in the updating process. In our view,
the propagation of labels is similar to information spreading. The more similar the two nodes are, the larger possibility that
they share the same opinion (label). In addition, the information is much easier to be propagated from
experts to common people, and not vice versa. That is to say, the label of an important node which may play a
central role in the network should be more likely to be retained in the updating process.
Here we assume that community centers are surrounded by neighbors with lower local densities and they
have a relatively low similarity with centers of other communities. Then the node influence can be used to guide the propagation.

If the influence of node $j$ to $i$, $\delta_{ij}$, is large, the mass given to the position that ``node $i$ adopts the labels of node $j$" should be
large. Suppose the set of neighbors of node $i$ is $N_i$, we then compute
  \begin{equation}
    \alpha_{ij} = \begin{cases}
      \varphi(\delta_{ij}) & \text{if}~j \in N_i,\\
      0 & \text{otherwise},
    \end{cases}
   \end{equation}
where $\varphi$ is a non-decreasing mapping from $[0, 1]$ to $[0,1]$.  We suggest to choose $\varphi$ as
  \begin{equation}\label{massdefine}
    \varphi(\delta_{ij}) = \alpha_0 \exp\bigg(-\gamma \frac{1-\delta_{ij}}{\delta_{ij}}\bigg),
  \end{equation}
where $\alpha_0$ and $\gamma$ are constants. Parameter $\alpha_0$ is a weight factor, and it can be set 1 by default.  Coefficient $\gamma$
can be fixed as follows \cite{denoeux2015ek}:
\begin{equation}
  \gamma = 1/\text{median}\left(\left\{\left(\frac{1-\delta_{ij}}{\delta_{ij}}\right)^2, i=1,2,\cdots,n, ~j \in N_i\right\}\right).
\end{equation}
   If node $j$ is a member of community $\omega_j$, then node $j$'s influence to node  $i$ is a piece of evidence that can
  be represented by the following mass function on $\Omega$:
  \begin{equation}
    m_j(\{\omega_j\}) = \alpha_{ij}, ~~
    m_j(\Omega) = 1 - \alpha_{ij}.
  \end{equation}
Let the number of elements in $N_i$ be $q_i$, and assume the influence from the $q_i$
nodes in the graph as independent pieces of evidence, the $q_i$ mass function $m_{j}$ can be then combined using the
DS rule:
\begin{equation}
  m = m_1 \oplus m_2 \oplus \cdots \oplus m_{q_i}.
\end{equation}

The fused mass $m$ is a credal membership of node $i$. The difference between this kind of membership and  fuzzy membership is that
there is a mass  assigned to the ignorant set $\Omega$ in bba $m$. It is used to describe the
probability that the node is an outlier of the graph.
The domain label of node $i$ can be defined as
\begin{equation}\label{domain1}
  Dl_i = \arg \max_{\omega_j}\{m(\{\omega_j\}), \omega_j \in \Omega\}.
\end{equation}
Since the focal elements of the bbas here are the singletons and set $\Omega$, Eq.~\eqref{domain1} is equal to
\begin{equation}
  Dl_i  = \arg \max_{\omega_j}\{pl(\omega_j), \omega_j \in \Omega\},
\end{equation}
where $pl$ is the contour function associative with $m$.

As explained in \cite{denoeux2015ek}, to obtain the domain label of each node, it is not necessary to compute the combined mass function $m$
explicitly.   For each node $i$, we first compute the logarithms of the plausibilities that node $i$ belongs to
cluster $\omega_k \in \Omega$ (up to an additive constant) as
\begin{equation}
  u_{ik}= \sum_{j \in N_i} v_{ij} s_{jk}, ~i = 1, 2, \cdots, N, j = 1, 2, \cdots, c
\end{equation}
where
\begin{equation}
  v_{ij} = -\log (1-\alpha_{ij}),
\end{equation}
and $s_{ik}$ is a logical variable indicating whether  the domain label of node $i$ is $\omega_k$.  Especially, if a
node has more than one dominant label, we
randomly choose a label from them as its dominant label.
The domain label of node $i$ can be set to $\omega_{k^{*}}$ with
\begin{equation}\label{assignlabel1}
  k^{*} = \arg \max_y \{u_{iy}\}.
\end{equation}
Then the variables $s_{ik}$ can be updated as
\begin{equation}\label{assignlabel2}
  s_{ik} = \begin{cases}
    1 & \text{if}~ k = k^{*}, \\
    0 & \text{otherwise}.
  \end{cases}
\end{equation}

The labels of each node are updated iteratively in ELP until the maximum iteration number is reached or all labels are stable.
Finally, the overlapping and non-overlapping communities are returned. The ELP algorithm  can be summarized in Algorithm \ref{alg:method}.

\begin{algorithm}\caption{\textbf{:}~~~ ELP  algorithm}\label{alg:method}
\begin{algorithmic}
\STATE {\textbf{Input:} Graph $G(V,E)$.}
\STATE{\textbf{Parameters:}
	~\\$\eta$: the parameter to adjust the node influence in Eq.~\eqref{siminf} \\
$T$: the maximum number of iteration steps\\
$\alpha_0, \gamma$: the parameters in Eq.~\eqref{massdefine} to define  mass functions }
\STATE {\textbf{Initialization:}\\
(1). Calculate the influence of node $j$ to node $i$, $\delta_{ij}$. \\
(2). Initialize a unique label of each node in the network. The matrix $\bm{S}= \{s_{ij}\}$ is initially set to be an identity matrix.
			}
\REPEAT
\STATE{
 (1). Arrange the nodes in the network  in a random order and save them in set $\sigma$ orderly. \\
 (2). Update the label of node $i$ one by one according to the order in $\sigma$.
 One can then assign node $i$ to the community  $\omega_k$ with the highest plausibility and update the
 matrix $\bm{S}$ using Eqs.~\eqref{assignlabel1} and \eqref{assignlabel2}.
}
\UNTIL{the maximum iteration number is reached or all domain labels become stable.}

\STATE{
\textbf{Output:} For each node, calculate the bba $m_i$ according to the labels of each node $i$, and output the bba matrix $\bm{M} = \{m_i\}$.
}
\end{algorithmic}
\end{algorithm}

\subsection{Update order}
In ELP, the labels of nodes are updated in a random order $\sigma$. Therefore, we may detect different communities
with different arrangements of nodes ({\em i.e.,} different  $\sigma$s), which leads to a stability concern.
Like LPA, ELP updates nodes' labels asynchronously. Benefiting from the asynchronous strategy, nodes which update
labels earlier with stable labels will have a positive impact on the ones updated later \cite{liu2015label}.

In order to find a good update order, we first introduce the concept of influence variance as
\begin{equation}
  V_i = \frac{\sum\limits_{j \in N_i}abs\left(\delta_{ij}^{*} - \frac{1}{|N_i|}\sum\limits_{t \in N_i}\delta^*_{it}\right)}{|N_i|},
\end{equation}
where $abs(x)$ is the absolute  value of $x$, $|N_i|$ is the number of elements in set $N_i$, and
\begin{equation}
  \delta_{ij} ^ * = \frac{\delta_{ij}}{\sum\limits_{t \in N_i} \delta_{it}}.
\end{equation}
We call $\delta_{ij}^*$ the normalized influence.

From the definition of influence variance, it can be seen that $V_i$ is small if the influence values of node $i$'s neighbors
do not spread out very much from the average. Hence,  we can conclude that if node $i$'s influence variance $V_{i}$
is large, there must be some neighbors with very large values of normalized influence. According to the label
propagation strategy,  the label of node $i$ is more likely to
updated to  the same one as the most influential neighbor. Therefore,
the larger the influence variance of a node is, the easier the node
updates its label.

In the label propagation strategy of ELP, the labels of central nodes will be easily adopted by  border nodes.
Thus if we set the labels of the border nodes as the same one with
the nearest centers first and the central nodes are updated later, the result of label propagation will be the same as the natural community of the network. The central nodes generally have large local densities. That is to day, the nodes with small local density should be updated first.

Based on the above analysis, in
order to identify the correct community structure, the nodes can be ordered based on $\beta$ index which can be defined as
\begin{equation}\label{updateorder}
  \beta_i =   \frac{v_i}{\sum_i v_i}+ \frac{\frac{1}{\rho_i}}{\sum_i \frac{1}{\rho_i}}.
\end{equation}
We arrange nodes in  a descending order of value $\beta$, and denote this order by $\sigma^*$.
\section{Experiments}
\label{secexp}
In this section, several experiments will be conducted on graphs. The results will be compared with LPA and  E$K$-NNclus. It should be noted  that
E$K$-NNclus is for relational data sets with given dissimilarities. To apply E$K$-NNclus on graph data sets, the following distance
measure associative with the similarity in Eq.~\eqref{siminf} is considered:
\begin{equation}
  d_{ij} = \frac{\delta_{ij}}{1 - \delta_{ij}}.
\end{equation}

We adopt the Normalized Mutual Information (NMI) \cite{danon2005comparing} to evaluate the quality of detected
communities.  The NMI of two partitions $A$ and $B$  of the graph, $\nmi(A,B)$, can be calculated  by
\begin{equation}
\label{nmi_eq}
 \nmi(A,B)=\frac{-2\sum_{i=1}^{C_A}\sum_{j=1}^{C_B}N_{ij}\log (\frac{N_{ij}n}{N_{i\cdot}N_{\cdot j}})}{
  \sum_{i=1}^{C_A}N_{i\cdot} \log(\frac{N_{i\cdot}}{n})+\sum_{j=1}^{C_B}N_{\cdot j} \log(\frac{N_{\cdot j}}{n})},
\end{equation}
where $C_A$ and $C_B$ denote the numbers of communities in partitions $A$ and $B$ respectively. The notation $N_{ij}$ denotes the element
of matrix $(\bm{N})_{C_A \times C_B}$, representing the number of nodes in the $i_{th}$ community of $A$ that appear
in the $j_{th}$ community of $B$. The sum over row $i$ of matrix $\bm{N}$  is denoted by $N_{i\cdot}$ and that over
column $j$ by $N_{\cdot j}$. If $A$ and $B$ are the same partitions, the NMI value is equal to one, {\em i.e.,} $$\nmi(A,B)=1.$$

\begin{center} \begin{figure}[!thbt] \centering
		\includegraphics[width=0.45\linewidth]{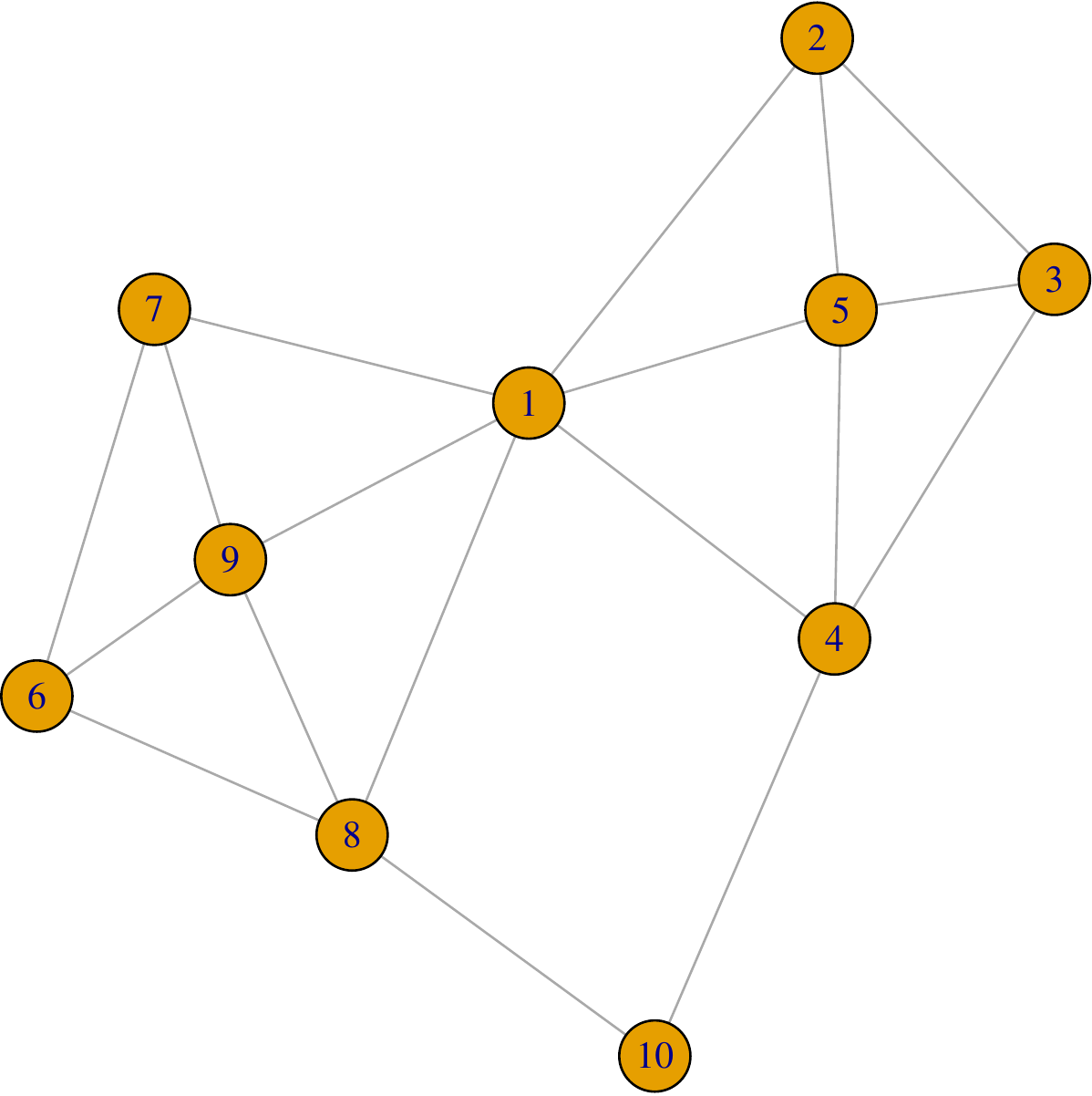}\hfill
        \includegraphics[width=0.45\linewidth]{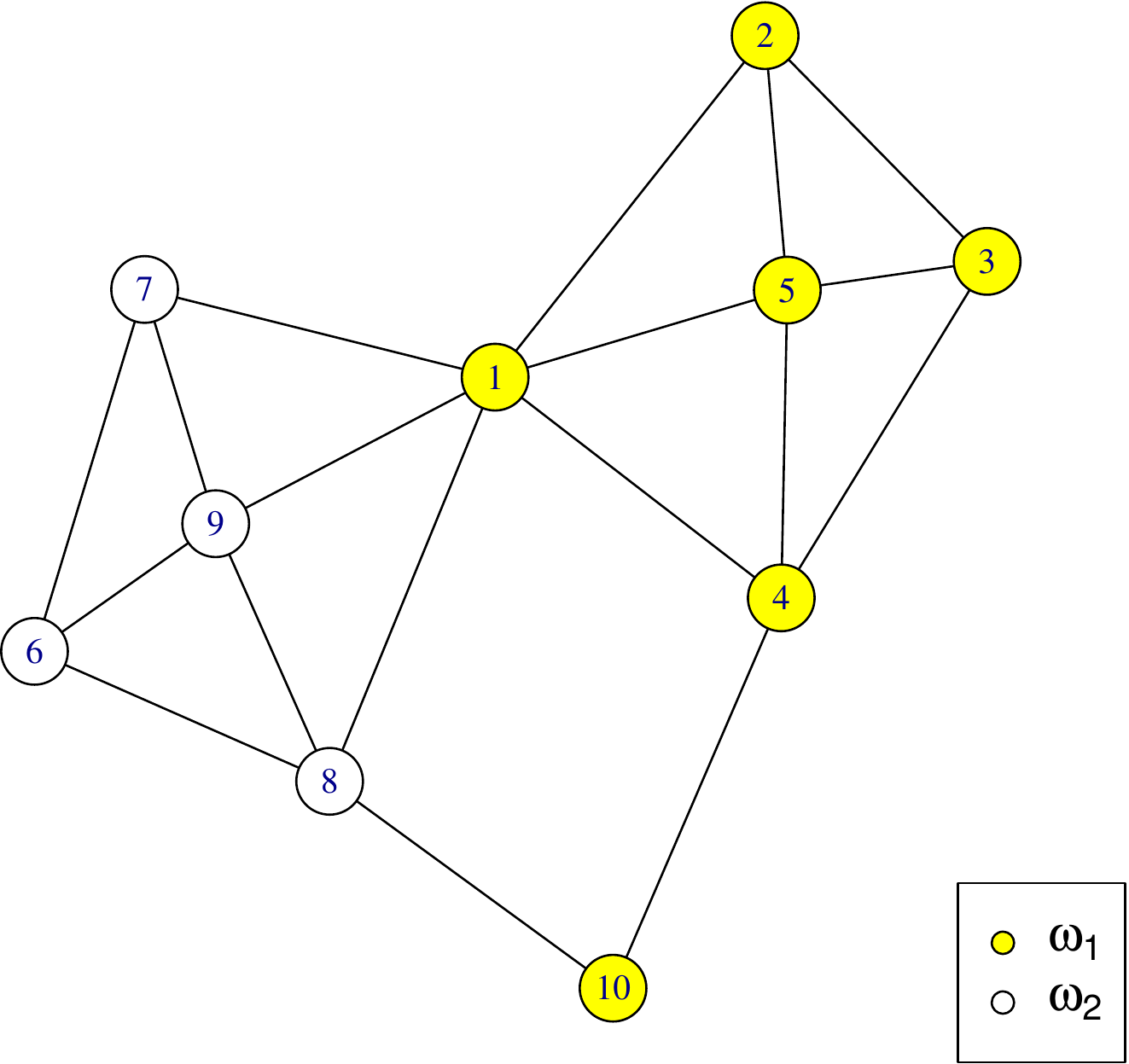} \hfill
        \parbox{.45\linewidth}{\centering\small a. Original network} \hfill
		\parbox{.45\linewidth}{\centering\small b. Detected communities by ELP with $\sigma^*$}
\caption{A network with an outlier and a bridge.} \label{example1} \end{figure} \end{center}

\Exa The network displayed in Fig. \ref{example1}--a contains ten nodes belonging to two communities.
Node 1 serves as a bridge between the nodes of two groups.

Let $\eta = 1$. We run ELP 50 times with different update order $\sigma$s. By partitioning
each node to the community with maximal plausibility value, the hard partition of all the nodes in the network can
be got. Nodes 2, 3, 4, 5 and nodes 6, 7, 8, 9 are correctly divided into two groups all the time. But the community labels
for nodes 1 and 10 are different using different update order. It indicates that it is difficult to determine the
specific labels of nodes 1 and 10 based on the simple topological graph structure.

\begin{center} \begin{figure}[!thbt] \centering
		\includegraphics[width=0.45\linewidth]{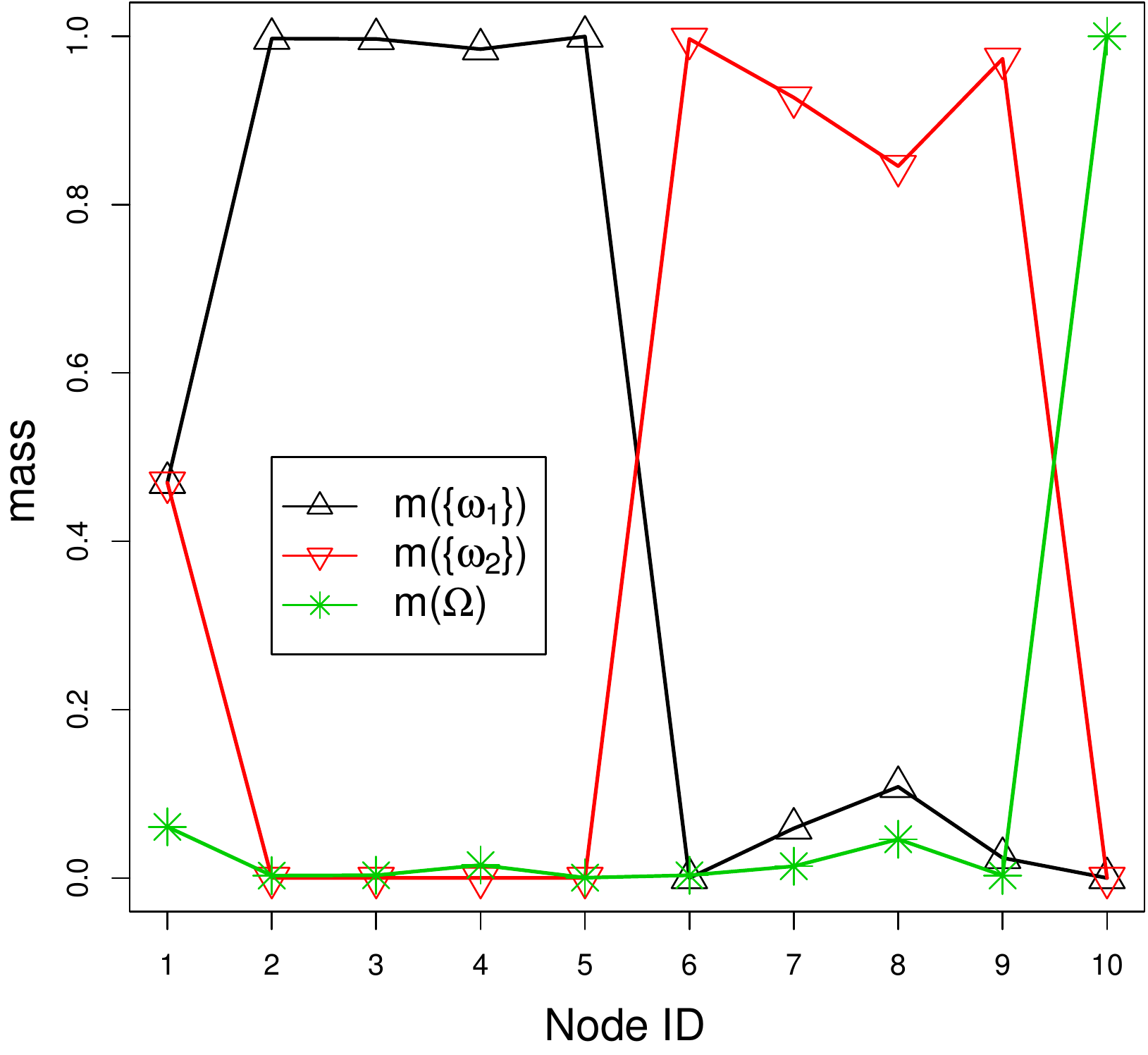}\hfill
      \includegraphics[width=0.45\linewidth]{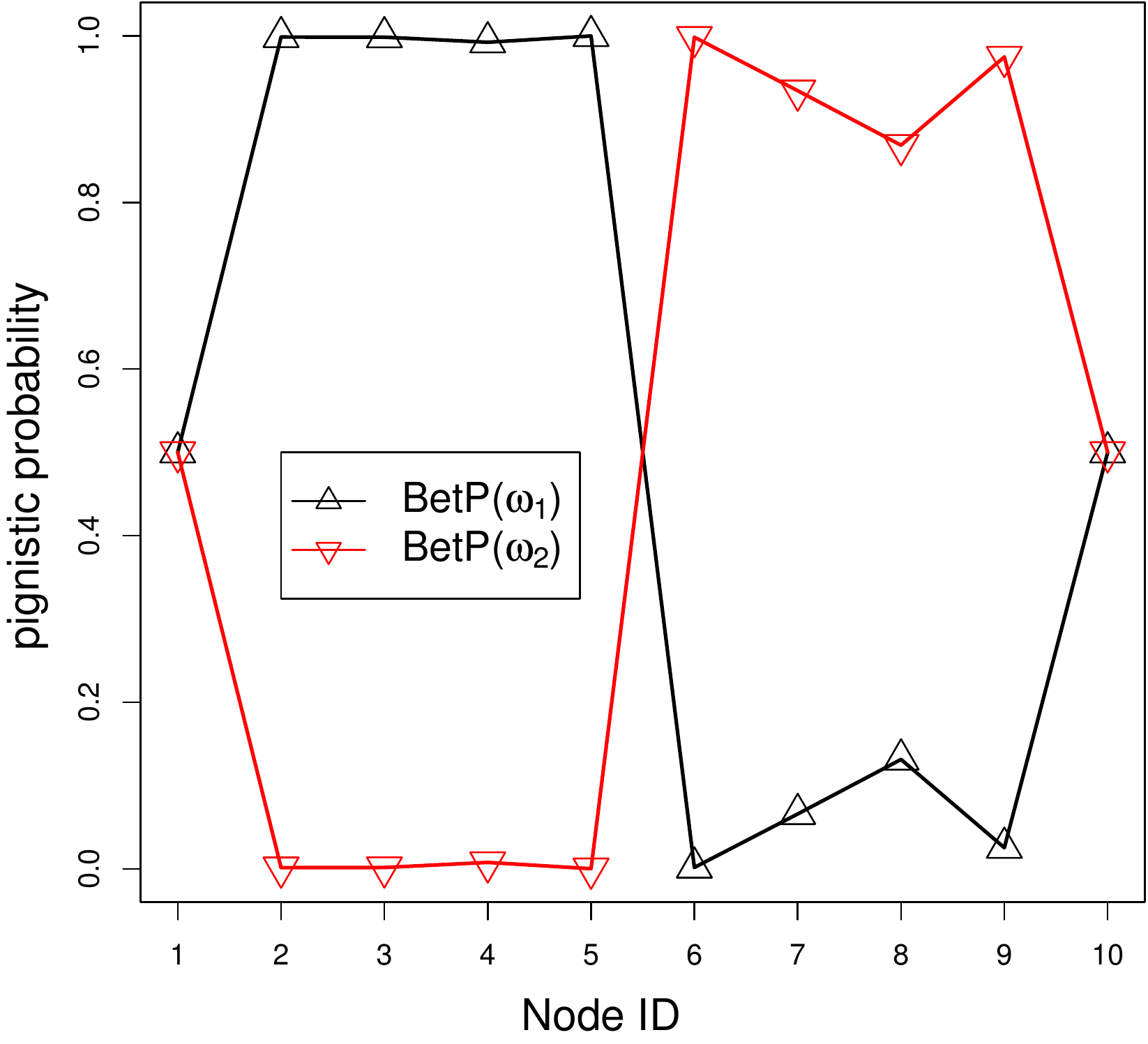} \hfill
        \parbox{.45\linewidth}{\centering\small a. Mass assignments} \hfill
		\parbox{.45\linewidth}{\centering\small b. Pignistic probabilities}
\caption{The bba and pignistic probabilities of each node by ELP.} \label{example1_bba} \end{figure} \end{center}

The optimal update order of nodes by Eq.~\eqref{updateorder} is  $$\sigma^*= \{4, 8,  3,  6,  2,  7,  1,  5,  9, 10\}.$$
The iterative update process of ELP with $\sigma^*$ is illustrated in Table \ref{elpupdate}. Initially, each node is
assigned  with a unique label which is the same as its ID. And then the nodes update their own labels orderly. The $i^\text{th}$ column  of the table
shows the label of each node after the $i^\text{th}$ update step. The bold element of each column in the table
indicates the node whose label  is updated in this step.  As can be seen, the
nodes located in the border are updated first. Based on the definition of node influence and the label propagation strategy, the
labels of central nodes are more easily adopted by the border nodes.
If the update order is set to be $\sigma^*$, the obtained hard partition
 is as that shown in Fig.~\ref{example1}--b. The corresponding  basic belief assignments for each node are
shown in Fig.~\ref{example1_bba}--a. As can be seen from the figure, the masses  of node 1 assigned
to the two communities are equal, indicating that node 1 serves as a bridge in the network.
For node 10, the maximum mass is given to the ignorant set $\Omega$.
Here $\Omega$ is the set for outliers which are very different from their neighbors. From the original graph, we can
see that node 10 has two neighbors,
nodes 4 and 8. But neither of them shares a common neighbor with node 10. Therefore, node 10 can be regarded as an outlier of the graph.
From the mass assignments, it is easy to see that the plausibilities of nodes 1 and 10 for two communities are equal. Consequently
it is difficult to determine their specific domain labels.

\begin{table*}[!htp]
\centering \caption{The iterative update process of ELP.}
\begin{tabular}{r|rrrrrrrrrrrrrrr}
  \hline
   \backslashbox{Node ID}{Step}        &          0 &          1 &          2 &          3 &          4 &          5 &          6 &          7 &          8 &          9 &         10 \\
\hline
         1 &          1 &          1 &          1 &          1 &          1 &          1 &          1 &    {\bf 5} &          5 &          5 &          5 \\

         2 &          2 &          2 &          2 &          2 &          2 &    {\bf 5} &          5 &          5 &          5 &          5 &          5 \\

         3 &          3 &          3 &          3 &    {\bf 5} &          5 &          5 &          5 &          5 &          5 &          5 &          5 \\

         4 &          4 &    {\bf 5} &          5 &          5 &          5 &          5 &          5 &          5 &          5 &          5 &          5 \\

         5 &          5 &          5 &          5 &          5 &          5 &          5 &          5 &          5 &    {\bf 5} &     5 &    {\bf 5} \\

         6 &          6 &          6 &          6 &          6 &    {\bf 9} &          9 &          9 &          9 &          9 &          9 &          9 \\

         7 &          7 &          7 &          7 &          7 &          7 &          7 &    {\bf 9} &          9 &          9 &          9 &          9 \\

         8 &          8 &          8 &    {\bf 9} &          9 &          9 &          9 &          9 &          9 &          9 &          9 &          9 \\

         9 &          9 &          9 &          9 &          9 &          9 &          9 &          9 &          9 &          9 &    {\bf 9} &          9 \\

        10 &         10 &         10 &         10 &         10 &         10 &         10 &         10 &         10 &         10 &         10 &          5 \\
        \hline
\end{tabular}\label{elpupdate}
\end{table*}

In ELP, pignistic probabilities can also be obtained as a by-product, which can be
regarded as fuzzy memberships of nodes.
From Fig.~\ref{example1_bba}--b we can see that both node 1 and node 10
have similar memberships for the two communities. But the positions of the two nodes in the graph are different. Node 1 is in the
central part while node 10 is in the  border. This is the
deficiency brought by the restriction that the probabilities over the frame
should be sum to 1. Consequently,
it could not distinguish  outliers from overlapping nodes. Although the mass values assigned
to the two communities are also equal, but those to set $\Omega$ are different. The
mass given to $\Omega$ for node 1 is  almost 0 while for node 10 it is approaching to 1.
It illustrates one of the advantages of ELP that  the overlapping nodes and outliers can be detected simultaneously
with the help of bbas. For one node, if the maximal mass is given to the ignorant
set $\Omega$, it is likely to be an outlier. On the contrary, when a node has large equal mass assignments for more than one
community, it  probably locates in the overlap.

\vspace{-2em}
We evoke LPA many times on this simple graph.  Nodes 1 and 10 are divided into
different communities in different runs. LPA could detect neither the overlapping nodes
nor the outliers. 
Before applying E$K$-NNclus algorithm, the number of nearest neighbors, $K$, should be fixed. The results by E$K$-NNclus with $K = 3$ and
$K =4$ are shown in Tables \ref{eknnclusk3} and \ref{eknnclusk4} respectively. It can be seen that five communities are detected
by E$K$-NNclus. Nodes 10 and  6 are specially partitioned into two special small groups respectively.
Node 1 is regarded as an outlier when $K=3$, while no outlier is detected when $K=4$.  In graphs,
different nodes have different number of neighbors, thus it is  not reasonable to use the same $K$ for all the nodes.
This may be the reason that the performance of E$K$-NNclus is not as good as that of ELP. The NMI values are not listed
in this experiment as there is no ground-truth for this illustrative graph.


\begin{table*}[!htp]
\centering \caption{The Mass Assignment by E$K$-NNclus with $K=3$.}
\begin{tabular}{r|rrrrrrrrrr}
  \hline
  \backslashbox{Mass}{Node ID} & 1 & 2 & 3 & 4 & 5 & 6 & 7 & 8 & 9 & 10 \\
  \hline
$m(\{\omega_1\})$ & 0.2600 & 0.0000 & \textbf{1.0000} & 0.0000 & 0.2588 & 0.0000 & 0.0000 & 0.1689 & \textbf{0.3261} & 0.0000 \\
  $m(\{\omega_2\})$ & 0.1098 & \textbf{1.0000} & 0.0000 & \textbf{0.8159} & \textbf{0.5767} & 0.0000 & 0.0000 & 0.1689 & 0.0000 & 0.0000 \\
  $m(\{\omega_3\})$ & 0.2600 & 0.0000 & 0.0000 & 0.0000 & 0.0000 & \textbf{1.0000} & 0.0000 & 0.0000 & 0.2091 & 0.0000 \\
  $m(\{\omega_4\})$ & 0.0000 & 0.0000 & 0.0000 & 0.1021 & 0.0000 & 0.0000 & \textbf{1.0000} & \textbf{0.5265} & 0.2576 & 0.0000 \\
  $m(\{\omega_5\})$ & 0.0000 & 0.0000 & 0.0000 & 0.0000 & 0.0000 & 0.0000 & 0.0000 & 0.0000 & 0.0000 & \textbf{1.0000} \\
  $m(\Omega)$ & \textbf{0.3702} & 0.0000 & 0.0000 & 0.0821 & 0.1644 & 0.0000 & 0.0000 & 0.1358 & 0.2072 & 0.0000 \\
   \hline
\end{tabular}\label{eknnclusk3}
\end{table*}

\begin{center} \begin{figure}[!thbt] \centering
		\includegraphics[width=0.6\linewidth]{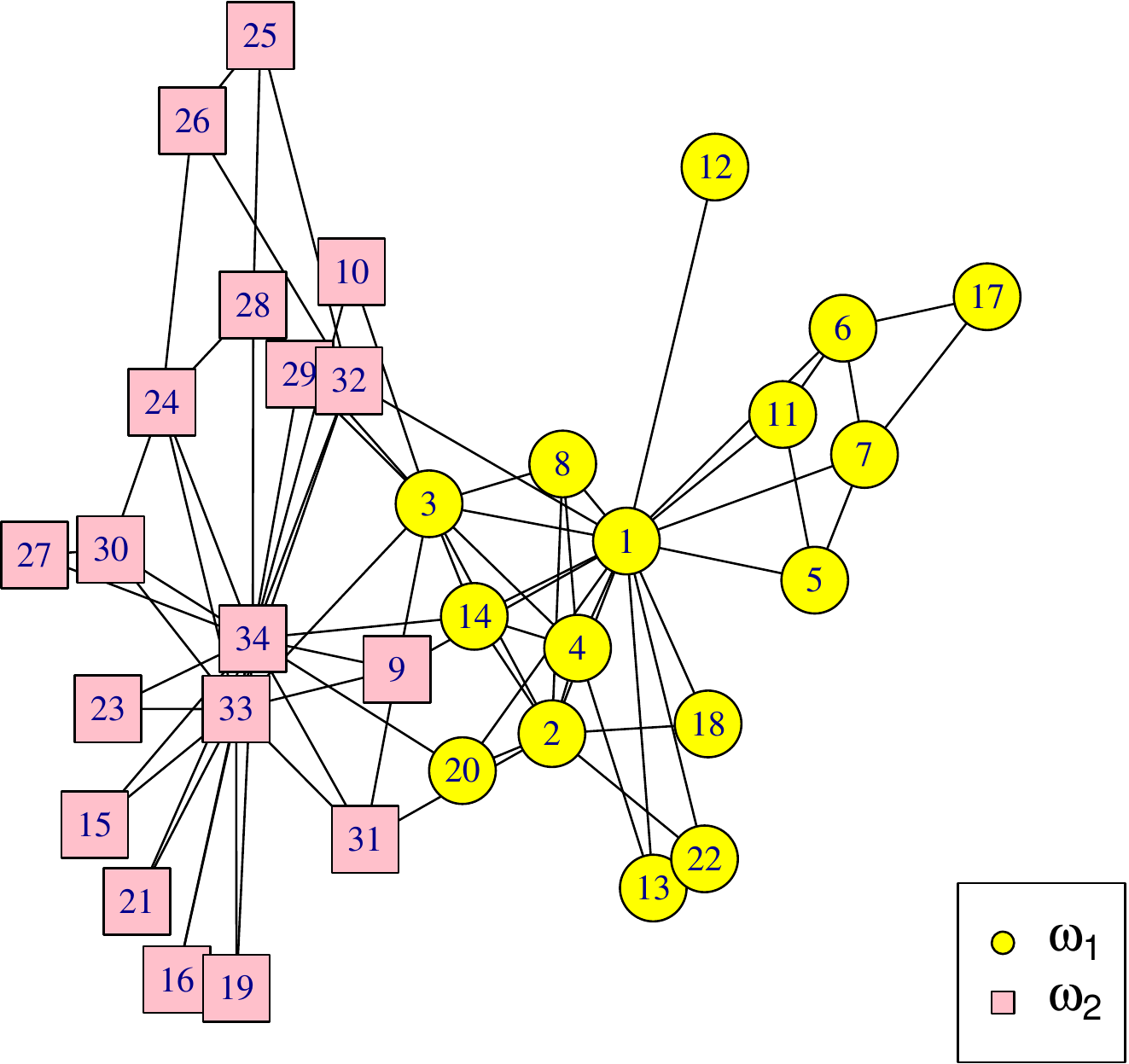}\hfill
\caption{Karate Club network.} \label{karate} \end{figure} \end{center}

\begin{table*}[!htp]
\centering \caption{The Mass Assignment by E$K$-NNclus with $K=4$.}
\begin{tabular}{r|rrrrrrrrrr}
  \hline
  \backslashbox{Mass}{Node ID} & 1 & 2 & 3 & 4 & 5 & 6 & 7 & 8 & 9 & 10 \\
  \hline
$m(\{\omega_1\})$ & \textbf{0.4190} & 0.0000 & \textbf{1.0000} & 0.2624 & 0.4951 & 0.0000 & 0.0000 & 0.0246 & 0.2162 & 0.0000 \\
  $m(\{\omega_2\})$ & 0.0000 & \textbf{1.0000} & 0.0000 & \textbf{0.4772} & \textbf{0.4171} & 0.0000 & 0.0000 & 0.0944 & 0.0000 & 0.0000 \\
  $m(\{\omega_3\})$ & 0.2265 & 0.0000 & 0.0000 & 0.0000 & 0.0000 & \textbf{1.0000} & 0.0000 & 0.0000 & 0.1411 & 0.0000 \\
  $m(\{\omega_4\})$ & 0.1016 & 0.0000 & 0.0000 & 0.1579 & 0.0000 & 0.0000 & \textbf{1.0000} & \textbf{0.8197} & \textbf{0.5309} & 0.0000 \\
  $m(\{\omega_5\})$ & 0.0000 & 0.0000 & 0.0000 & 0.0000 & 0.0000 & 0.0000 & 0.0000 & 0.0000 & 0.0000 & \textbf{1.0000} \\
  $m(\Omega)$ & 0.2530 & 0.0000 & 0.0000 & 0.1024 & 0.0878 & 0.0000 & 0.0000 & 0.0613 & 0.1118 & 0.0000 \\
   \hline
\end{tabular}\label{eknnclusk4}
\end{table*}
\Exa Here we  test on a widely used benchmark  in detecting community structures, ``Karate Club",  studied by
Wayne Zachary \cite{ucidatasets}. The network consists of 34  nodes and 78 edges representing the friendship among the members of
the club (see Fig. \ref{karate}).
During the development, a dispute arose between the club's administrator and instructor, which eventually resulted in the club  split into two smaller
clubs (one marked with pink squares,
and the other marked with yellow circles), centered around the administrator and the instructor respectively.

\begin{center} \begin{figure}[!thbt] \centering
		\includegraphics[width=0.45\linewidth]{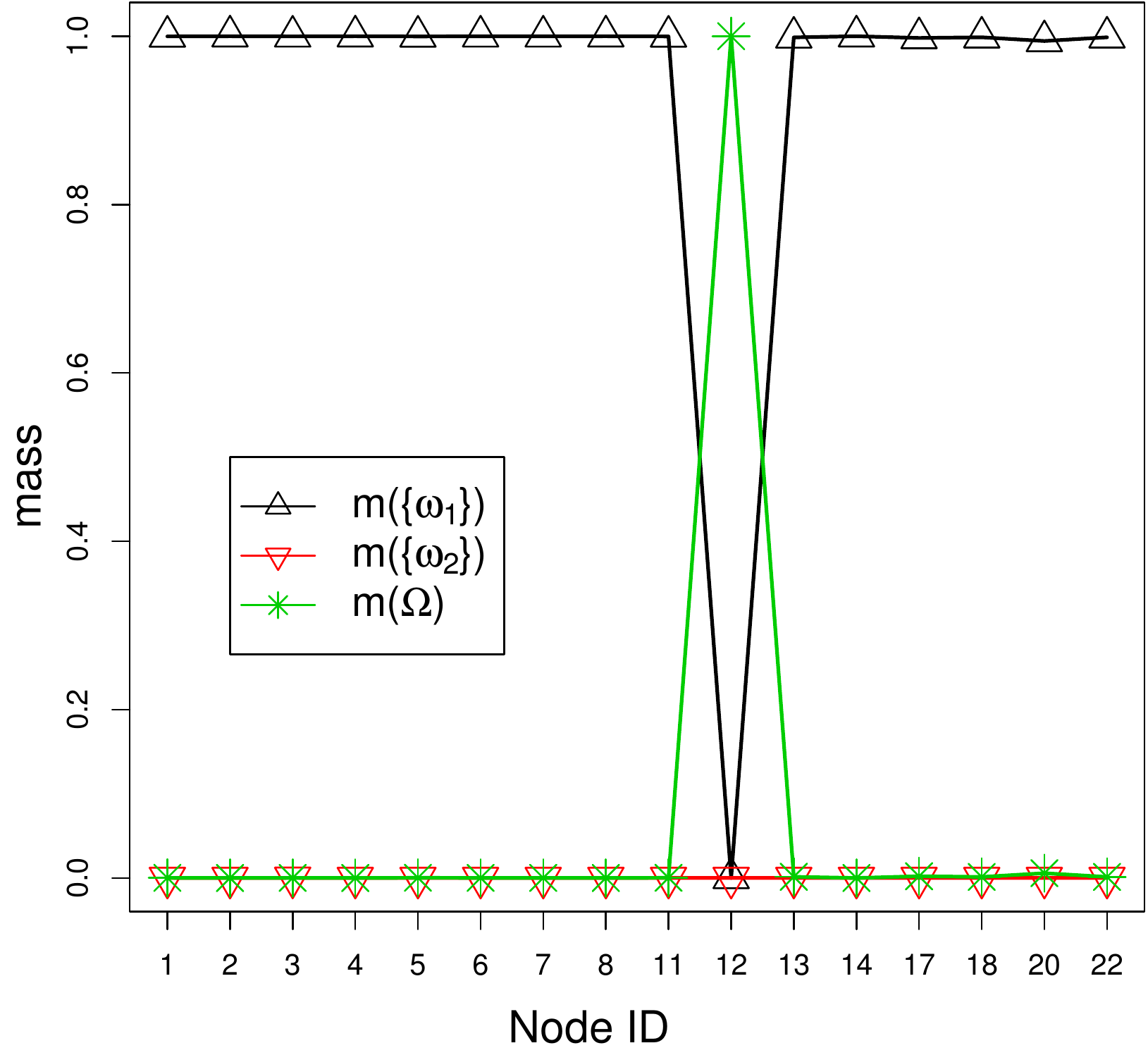}\hfill
      \includegraphics[width=0.45\linewidth]{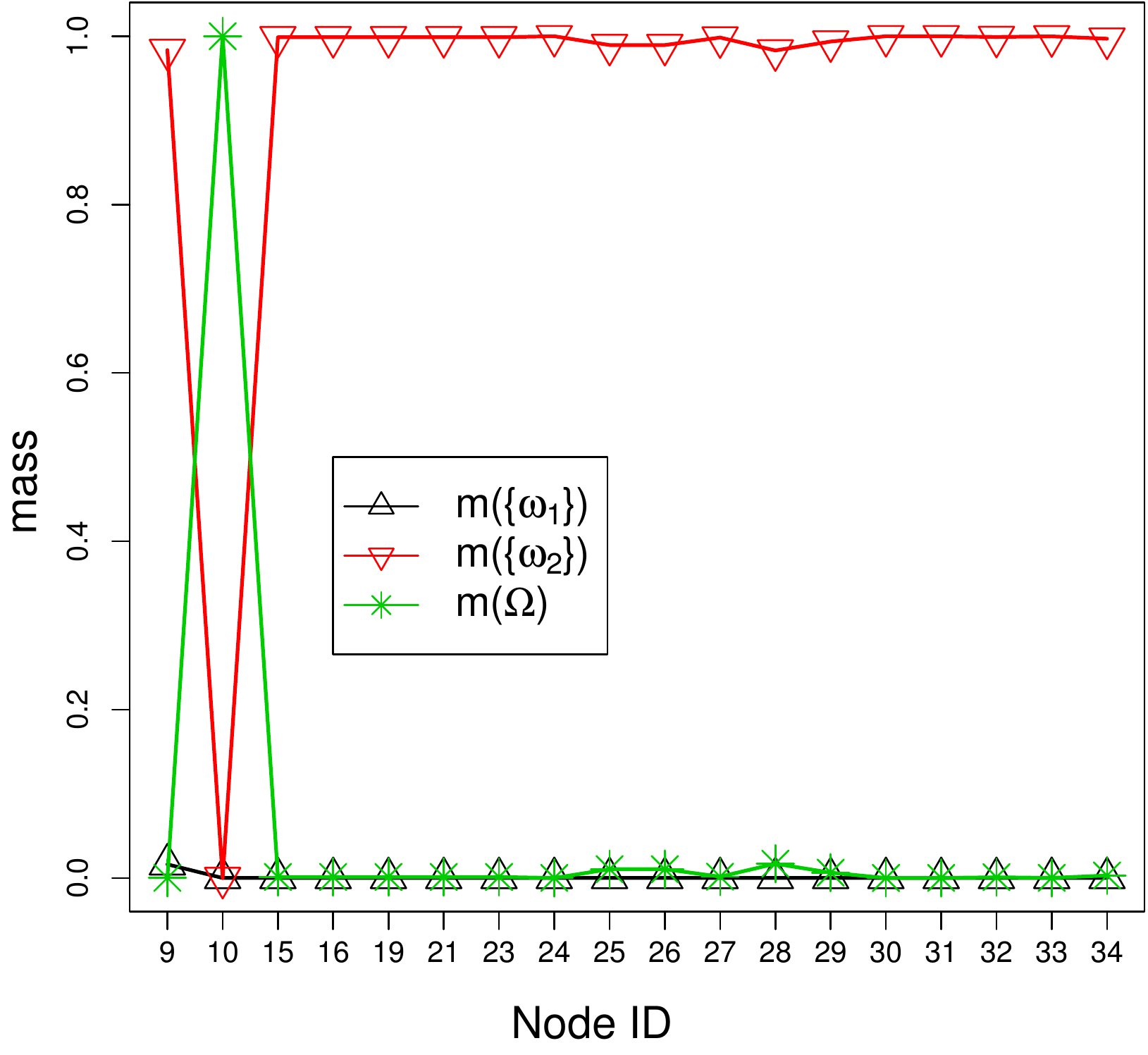} \hfill
        \parbox{.45\linewidth}{\centering\small a. Community $\omega_1$} \hfill
		\parbox{.45\linewidth}{\centering\small b. Community $\omega_2$}
\caption{The bba of each node for Karate Club network by ELP.} \label{karate_bba} \end{figure} \end{center}

\vspace{-1em}
Let $\eta = 1$, and evoke ELP many times with different update order $\sigma$s. We find that we can get two different results. Most of the time,
ELP could detect two communities and find two outliers. The bbas of nodes in the two groups are illustrated
in Figs.~\ref{karate_bba}--a and \ref{karate_bba}--b respectively.  It is showed in the
figures that this network has strong class structures, since for
each node the mass values assigned to different classes are significantly different. Nodes 10 and  12 are two outliers in their own communities.
From the original graph, node 12 only connects with node 1. For node 10, it has two neighbors, nodes 3 and 34, but it has no connection with the
neighbors of the two nodes. Neither node 10 nor node 12 has  close relationship with
other nodes in the network. Therefore, it is very intuitive  that they are detected as outliers. It is noted here that with update
order $\sigma^*$, we can get the above clustering result with two communities and two outliers.
\begin{center} \begin{figure}[!thbt] \centering
		\includegraphics[width=0.45\linewidth]{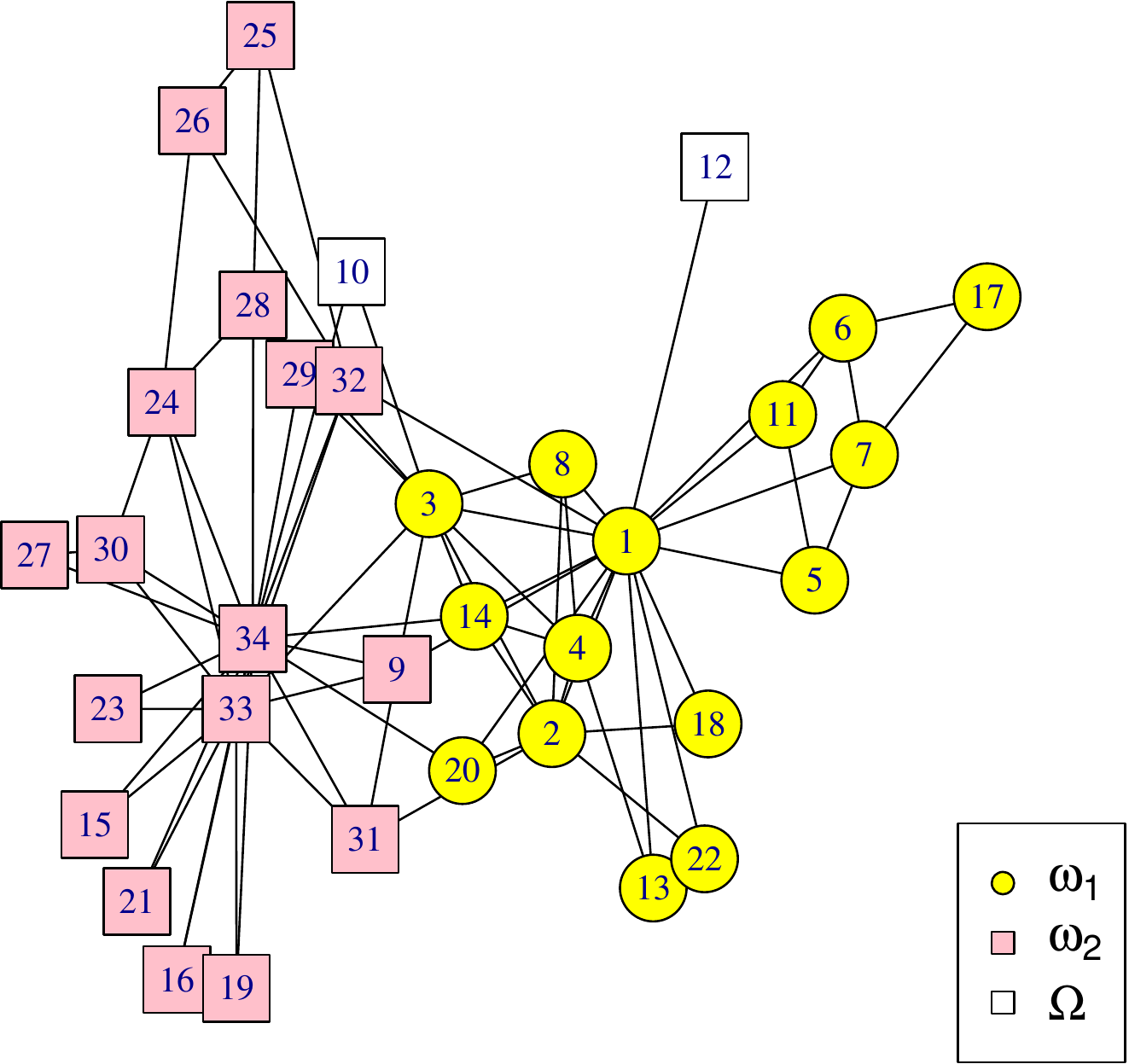}\hfill
     \includegraphics[width=0.45\linewidth]{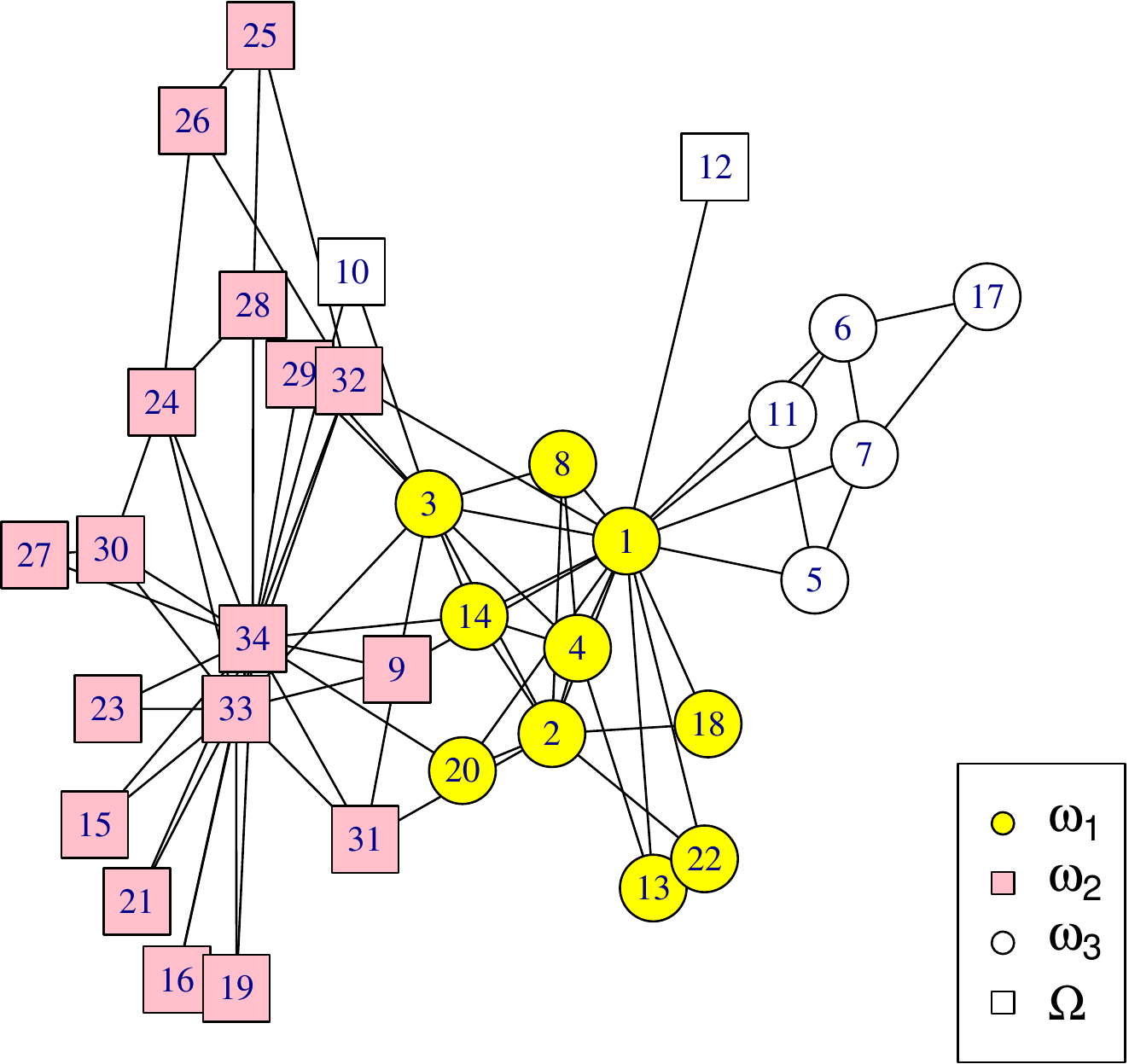}\hfill
     \parbox{.45\linewidth}{\centering\small a. Two detected communities}
        \parbox{.45\linewidth}{\centering\small b. Three detected communities}
\caption{The detected communities of Karate Club network by ELP.} \label{karateELP} \end{figure} \end{center}

\begin{center} \begin{figure}[!thbt] \centering
		\includegraphics[width=0.6\linewidth]{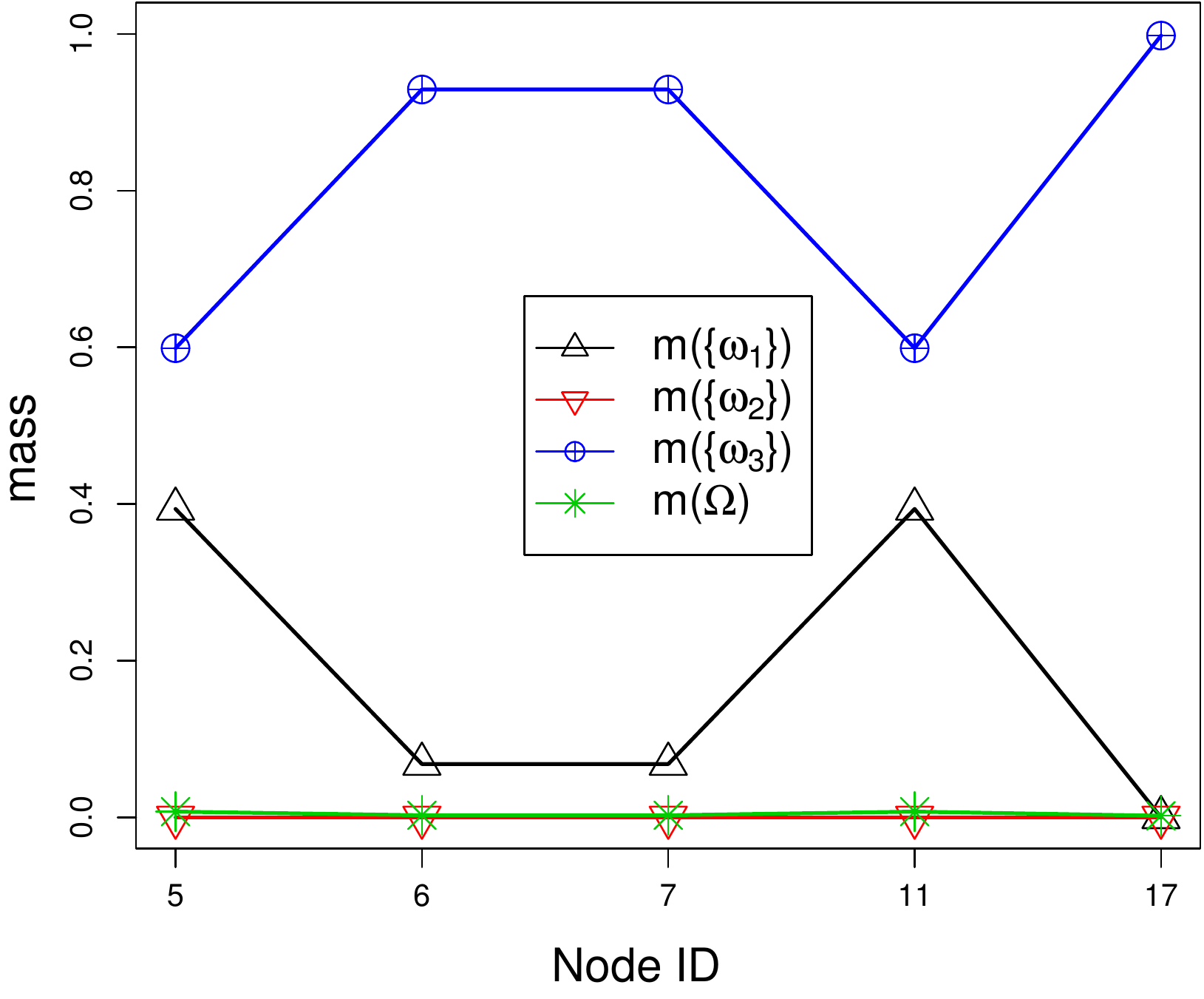}\hfill
\caption{The bbas of the nodes in small communities Karate Club network.} \label{karate_cluster3_bba} \end{figure} \end{center}

\vspace{-4em}
With some $\sigma$s, a special small community can been
found by ELP. As shown in Fig.~\ref{karateELP}--b, a group containing nodes 5, 6, 7, 11, 17 has been
separated from community $\omega_1$. This seems  reasonable as these nodes have no connections
with other nodes in class $\omega_1$ except the central node (node 1). The bbas of these five
nodes are illustrated in Fig.~\ref{karate_cluster3_bba}. It can be seen that nodes 5 and 11 have
large mass values for community $\omega_1$ and $\omega_3$, which can be regarded as bridges of
the two communities. Nodes 10 and 12 are still regarded as outliers in this case.

\vspace{-2em}
To compare the accuracy and robustness of different methods, each algorithm (ELP, LPA and E$K$-NNclus with $K$ = 3, 4, 5, 6) is repeated 50 times
with a random update order each time.
The minimum, maximum, average and the standard deviation of NMI values are listed in Table \ref{karate_nmi}. To get NMI values of the detected results
of ELP, we should get the domain label of each node by assigning each node
to the community with maximum plausibility. It should be noted that node 10 has a equal plausibility for
community $\omega_1$ and $\omega_2$, that is,
$$pl(\omega_1) = pl(\omega_2).$$
We randomly set a label as its dominant label. From the table we can find that ELP has good robustness as well as
accuracy.  E$K$-NNclus is stable when $K$ is relatively large, but the accuracy is not as good as that in ELP. The
performance of LPA
is worst in terms of  stability and average accuracy.
\begin{table}[ht]
\centering \caption{The NMI values for the detected results by different methods on Karate Club network.}
\begin{tabular}{rrrrrrrr}
  \hline
&      &     & \multicolumn{4}{c}{E$K$-NNclus} \\
\cline{4-7}
 & ELP & LPA &$K=3$ & $K=4$ & $K=5$ & $K=6$\\
  \hline
  Max & 1.0000 & 1.0000 & 0.4648 & 0.4832 & 0.5248 & 0.5248 \\
  Min & 0.8255 & 0.0000 & 0.4149 & 0.4149 & 0.5248 & 0.5248 \\
  Average & 0.9314 & 0.6679 & 0.4498 & 0.4711 & 0.5248 & 0.5248 \\
  Deviation & 0.0815 & 0.1945 & 0.0231 & 0.0262 & 0.0000 & 0.0000 \\
   \hline
\end{tabular}\label{karate_nmi}
\end{table}

\begin{center} \begin{figure}[!thbt] \centering
		\includegraphics[width=0.45\linewidth]{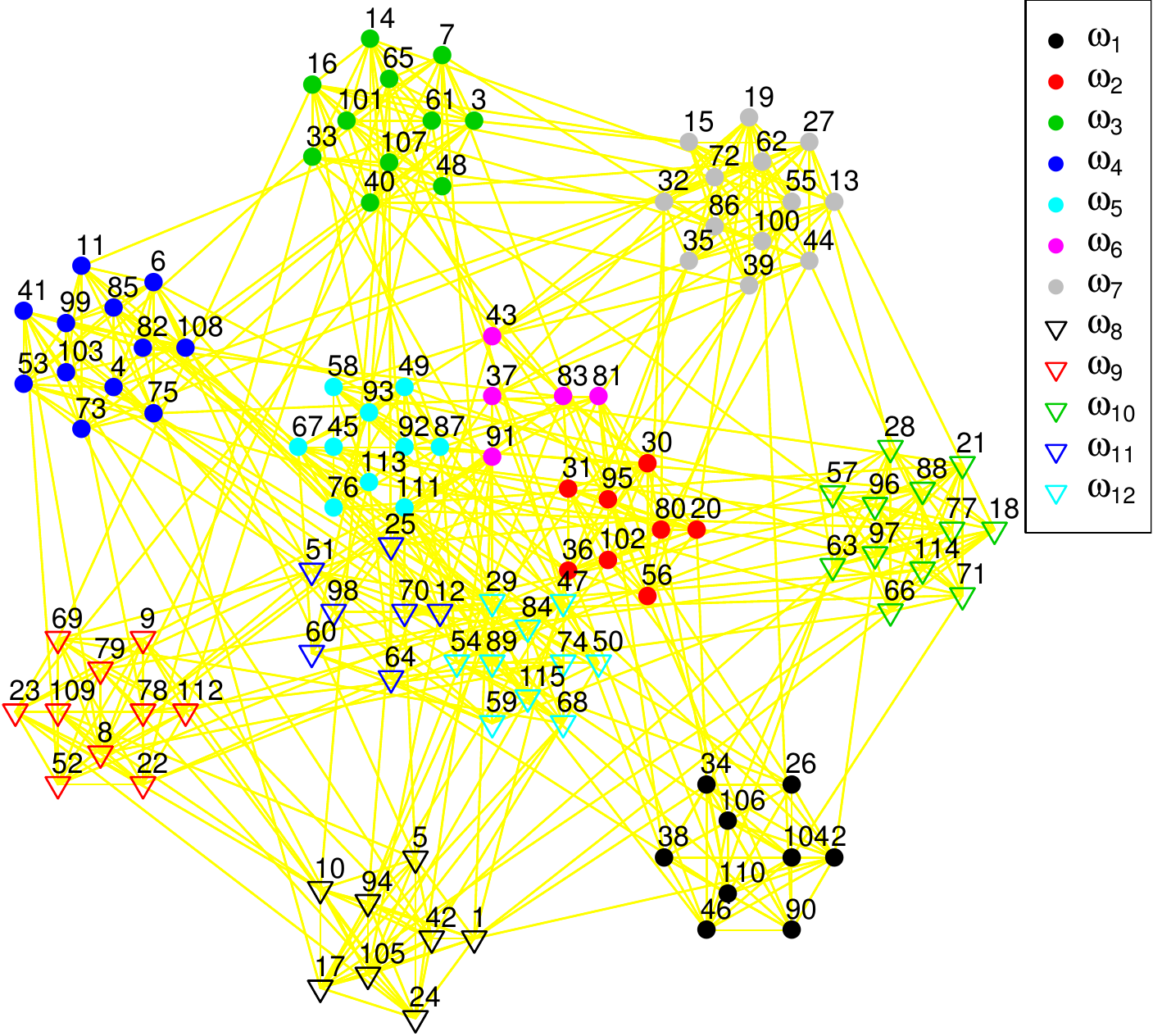}\hfill
     \includegraphics[width=0.45\linewidth]{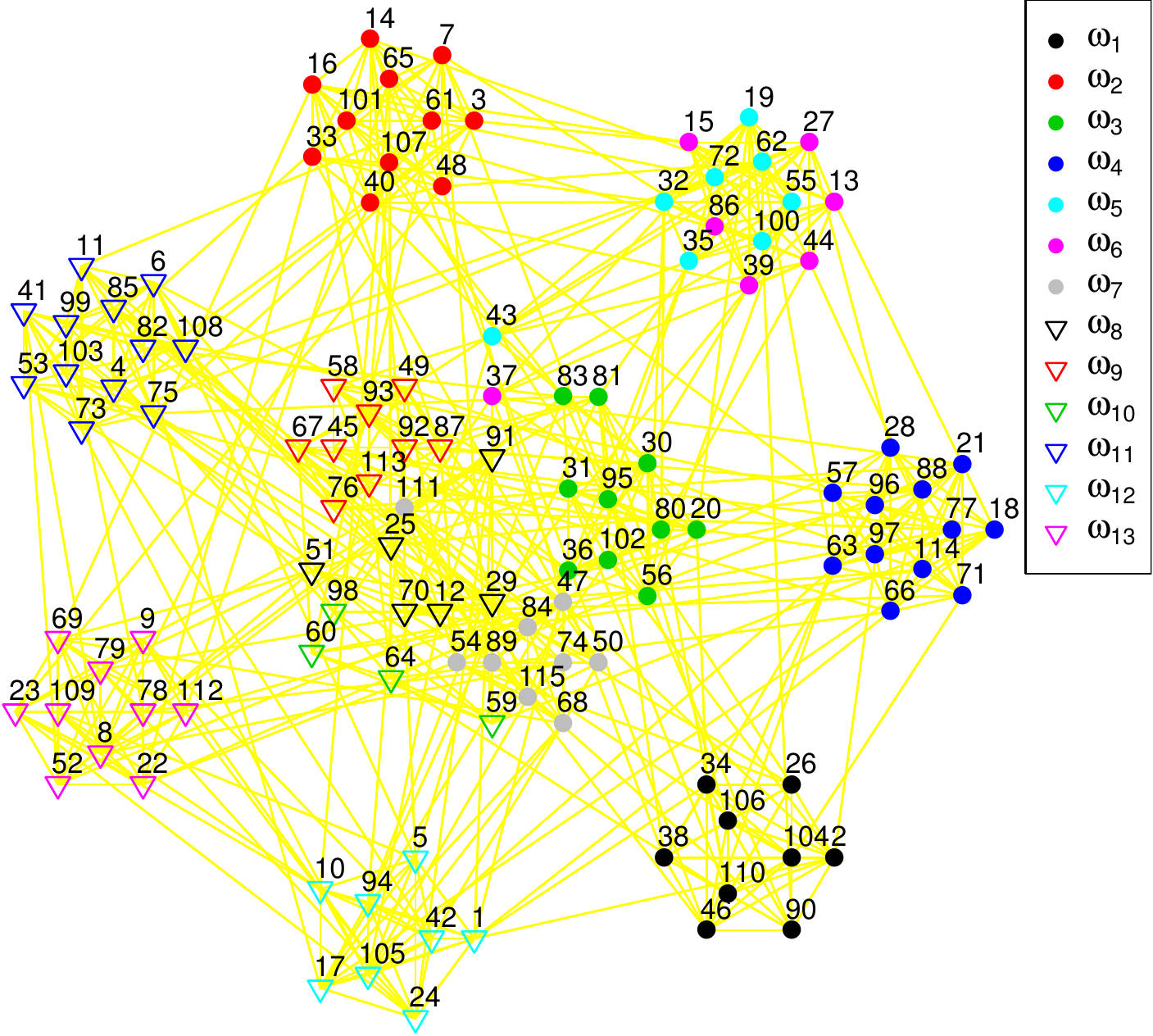}\hfill
     \parbox{.45\linewidth}{\centering\small a. Original network}
        \parbox{.45\linewidth}{\centering\small b. The detected communities}
\caption{American football network.} \label{footballELP} \end{figure} \end{center}

\Exa The network we investigate in
this experiment is the world of American college football games between
Division IA colleges during regular season Fall 2000. The vertices  in the network represent 115 teams,
while the links denote 613 regular-season games between the two teams they
connect. The teams are divided into 12 conferences  containing around 8-12
teams each and generally games are more frequent between members from the same
conference than between those from different conferences. The original network is displayed in Fig.~\ref{footballELP}--a.

Each of the algorithms ELP, LPA, and E$K$-NNclus is repeated 50 times with different update order $\sigma$s, the statistical properties of the
corresponding NMI values are listed in Table \ref{football_nmi}. As can be seen, the maximal value of NMI is almost the same by LPA and ELP.
However, the minimum and average by ELP are significantly larger  than those by LPA. These results further demonstrate the robustness of ELP.

\begin{table}[ht]
\centering \caption{The NMI values for the detected results by different methods on Football network.}
\begin{tabular}{rrrrrrrr}
  \hline
&      &     & \multicolumn{4}{c}{E$K$-NNclus} \\
\cline{4-7}
 & ELP & LPA &$K=3$ & $K=4$ & $K=5$ & $K=6$\\
  \hline
Max & 0.9269 & 0.9269 & 0.8376 & 0.8730 & 0.9030 & 0.9030 \\
  Min & 0.8892 & 0.8343 & 0.8024 & 0.8103 & 0.8404 & 0.8688 \\
  Average & 0.9061 & 0.8887 & 0.8166 & 0.8384 & 0.8700 & 0.8860 \\
  Deviation & 0.0080 & 0.0232 & 0.0104 & 0.0141 & 0.0139 & 0.0092 \\
   \hline
\end{tabular}\label{football_nmi}
\end{table}

Now we fix the update order and let $\sigma = \sigma^*$. The NMI value of the
detected communities is 0.9102. It is very close to the maximum 0.9269.  The  clustering result of
ELP with $\sigma^*$ is presented in Fig.~\ref{footballELP}--b. As shown in the figure, six conferences
are exactly identified.

\vspace{-2em}
\Exa In this experiment, we will test  on three other real-world graphs: Dolphins network, Lesmis network and Political
books network \cite{ucidatasets}. 
For one data set, each
algorithm is evoked 50 times with a random $\sigma$ each time.
The  statistical characteristics of the evaluation results in terms of NMI on the three
data sets are illustrated in Tables \ref{dolphin_nmi} -- \ref{books_nmi} respectively. As  shown in the tables,
ELP is much more stable than other approaches as the standard deviation is quite small. The average performance of ELP is best among all the methods.

\begin{table}[ht]
\centering \caption{The NMI values for the detected results by different methods on Dolphin network.}
\begin{tabular}{rrrrrrrr}
  \hline
&      &     & \multicolumn{4}{c}{E$K$-NNclus} \\
\cline{4-7}
 & ELP & LPA &$K=3$ & $K=4$ & $K=5$ & $K=6$\\
  \hline
Max & 0.8230 & 1.0000 & 0.4975 & 0.5400 & 0.6729 & 0.6089 \\
  Min & 0.5815 & 0.4689 & 0.4835 & 0.4786 & 0.5371 & 0.4774 \\
  Average & 0.6346 & 0.6450 & 0.4964 & 0.5034 & 0.5834 & 0.5268 \\
  Deviation & 0.0504 & 0.1113 & 0.0038 & 0.0205 & 0.0549 & 0.0541 \\
     \hline
\end{tabular}\label{dolphin_nmi}
\end{table}

\begin{table}[ht]
\centering \caption{The NMI values for the detected results by different methods on Lesmis network.}
\begin{tabular}{rrrrrrrr}
  \hline
&      &     & \multicolumn{4}{c}{E$K$-NNclus} \\
\cline{4-7}
 & ELP & LPA &$K=3$ & $K=4$ & $K=5$ & $K=6$\\
  \hline
Max & 0.8645 & 0.8441 & 0.1475 & 0.1475 & 0.5357 & 0.5357 \\
  Min & 0.8645 & 0.5114 & 0.1055 & 0.1055 & 0.4153 & 0.4254 \\
  Average & 0.8645 & 0.6907 & 0.1190 & 0.1374 & 0.4963 & 0.5122 \\
  Deviation & 0.0000 & 0.0705 & 0.0198 & 0.0181 & 0.0429 & 0.0306 \\
   \hline
\end{tabular}\label{lemis_nmi}
\end{table}

\begin{table}[ht]
\centering \caption{The NMI values for the detected results by different methods on Books network.}
\begin{tabular}{rrrrrrrr}
  \hline
&      &     & \multicolumn{4}{c}{E$K$-NNclus} \\
\cline{4-7}
 & ELP & LPA &$K=3$ & $K=4$ & $K=5$ & $K=6$\\
  \hline
Max & 0.5751 & 0.5979 & 0.4348 & 0.4111 & 0.4421 & 0.4563 \\
  Min & 0.4979 & 0.4607 & 0.4348 & 0.4111 & 0.4421 & 0.4563 \\
  Average & 0.5496 & 0.5535 & 0.4348 & 0.4111 & 0.4421 & 0.4563 \\
  Deviation & 0.0129 & 0.0333 & 0.0000 & 0.0000 & 0.0000 & 0.0000 \\
   \hline
\end{tabular}\label{books_nmi}
\end{table}


\section{Conclusion}
In this paper, a new community detection approach, named ELP, is presented.  The proposed approach is inspired from the
conventional LPA and E$K$-NNclus clustering algorithm.  By the introduction of node influence,
a new evidential label propagation strategy is devised. After the  propagation process, the domain label of each node
is determined according to its plausibilities.
The experimental results illustrate the advantages of ELP. It can
be used to detect the overlapping nodes and outliers at the same time. To define the influence of each node,
different similarity measures can be adopted. Specially, if there are some attributes describing the features of nodes,
a similarity index considering both the topological graph structure and the attribute information is a better choice.
Therefore, we intend to discuss the effects of different similarity measures on ELP and the application of ELP on graphs
with attribute information   in our future research work.
\section*{Acknowledgements}
This work was supported by the National
Natural Science Foundation of China (Nos.61135001, 61403310).

\bibliographystyle{IEEEtran}
\bibliography{paperlist}

\begin{thebibliography}{10}
\providecommand{\url}[1]{#1}
\csname url@samestyle\endcsname
\providecommand{\newblock}{\relax}
\providecommand{\bibinfo}[2]{#2}
\providecommand{\BIBentrySTDinterwordspacing}{\spaceskip=0pt\relax}
\providecommand{\BIBentryALTinterwordstretchfactor}{4}
\providecommand{\BIBentryALTinterwordspacing}{\spaceskip=\fontdimen2\font plus
\BIBentryALTinterwordstretchfactor\fontdimen3\font minus
  \fontdimen4\font\relax}
\providecommand{\BIBforeignlanguage}[2]{{%
\expandafter\ifx\csname l@#1\endcsname\relax
\typeout{** WARNING: IEEEtran.bst: No hyphenation pattern has been}%
\typeout{** loaded for the language `#1'. Using the pattern for}%
\typeout{** the default language instead.}%
\else
\language=\csname l@#1\endcsname
\fi
#2}}
\providecommand{\BIBdecl}{\relax}
\BIBdecl

\bibitem{newman2004finding}
M.~E. Newman and M.~Girvan, ``Finding and evaluating community structure in
  networks,'' \emph{Physical review E}, vol.~69, no.~2, p. 026113, 2004.

\bibitem{fortunato2007resolution}
S.~Fortunato and M.~Barthelemy, ``Resolution limit in community detection,''
  \emph{Proceedings of the National Academy of Sciences}, vol. 104, no.~1, pp.
  36--41, 2007.

\bibitem{kim2015detecting}
P.~Kim and S.~Kim, ``Detecting overlapping and hierarchical communities in
  complex network using interaction-based edge clustering,'' \emph{Physica A:
  Statistical Mechanics and its Applications}, vol. 417, pp. 46--56, 2015.

\bibitem{newman2013spectral}
M.~E. Newman, ``Spectral methods for community detection and graph
  partitioning,'' \emph{Physical Review E}, vol.~88, no.~4, p. 042822, 2013.

\bibitem{zhou2015similarity}
K.~Zhou, A.~Martin, and Q.~Pan, ``A similarity-based community detection method
  with multiple prototype representation,'' \emph{Physica A: Statistical
  Mechanics and its Applications}, vol. 438, pp. 519--531, 2015.

\bibitem{raghavan2007near}
U.~N. Raghavan, R.~Albert, and S.~Kumara, ``Near linear time algorithm to
  detect community structures in large-scale networks,'' \emph{Physical Review
  E}, vol.~76, no.~3, p. 036106, 2007.

\bibitem{ds2}
G.~Shafer, \emph{A mathematical theory of evidence}.\hskip 1em plus 0.5em minus
  0.4em\relax Princeton University Press, 1976.

\bibitem{denoeux1995k}
T.~Den{\oe}ux, ``A $k$-nearest neighbor classification rule based on
  dempster-shafer theory,'' \emph{Systems, Man and Cybernetics, IEEE
  Transactions on}, vol.~25, no.~5, pp. 804--813, 1995.

\bibitem{liu2015new}
Z.-g. Liu, Q.~Pan, G.~Mercier, and J.~Dezert, ``A new incomplete pattern
  classification method based on evidential reasoning,'' \emph{Cybernetics,
  IEEE Transactions on}, vol.~45, no.~4, pp. 635--646, 2015.

\bibitem{masson2008ecm}
M.-H. Masson and T.~Den{\oe}ux, ``{ECM}: An evidential version of the fuzzy
  $c$-means algorithm,'' \emph{Pattern Recognition}, vol.~41, no.~4, pp.
  1384--1397, 2008.

\bibitem{zhou2015evidential}
K.~Zhou, A.~Martin, Q.~Pan, and Z.-G. Liu, ``Evidential relational clustering
  using medoids,'' in \emph{Information Fusion (Fusion), 2015 18th
  International Conference on}.\hskip 1em plus 0.5em minus 0.4em\relax IEEE,
  2015, pp. 413--420.

\bibitem{zhou2016ecmdd}
K.~Zhou, A.~Martin, Q.~Pan, and Z.-g. Liu, ``{ECM}dd: Evidential $c$-medoids
  clustering with multiple prototypes,'' \emph{Pattern Recognition}, doi:
  10.1016/j.patcog.2016.05.005, 2016.

\bibitem{wei2013identifying}
D.~Wei, X.~Deng, X.~Zhang, Y.~Deng, and S.~Mahadevan, ``Identifying influential
  nodes in weighted networks based on evidence theory,'' \emph{Physica A:
  Statistical Mechanics and its Applications}, vol. 392, no.~10, pp.
  2564--2575, 2013.

\bibitem{zhou2014evidential}
K.~Zhou, A.~Martin, and Q.~Pan, ``Evidential communities for complex
  networks,'' in \emph{Information Processing and Management of Uncertainty in
  Knowledge-Based Systems}.\hskip 1em plus 0.5em minus 0.4em\relax Springer,
  2014, pp. 557--566.

\bibitem{zhou2015median}
K.~Zhou, A.~Martin, Q.~Pan, and Z.-g. Liu, ``Median evidential $c$-means
  algorithm and its application to community detection,'' \emph{Knowledge-Based
  Systems}, vol.~74, pp. 69--88, 2015.

\bibitem{denoeux2013maximum}
T.~Den{\oe}ux, ``Maximum likelihood estimation from uncertain data in the
  belief function framework,'' \emph{Knowledge and Data Engineering, IEEE
  Transactions on}, vol.~25, no.~1, pp. 119--130, 2013.

\bibitem{zhou2014evidentialem}
K.~Zhou, A.~Martin, and Q.~Pan, ``Evidential-{EM} algorithm applied to
  progressively censored observations,'' in \emph{Information Processing and
  Management of Uncertainty in Knowledge-Based Systems}.\hskip 1em plus 0.5em
  minus 0.4em\relax Springer, 2014, pp. 180--189.

\bibitem{smets2005decision}
P.~Smets, ``Decision making in the {TBM}: the necessity of the pignistic
  transformation,'' \emph{International Journal of Approximate Reasoning},
  vol.~38, no.~2, pp. 133--147, 2005.

\bibitem{denoeux2015ek}
T.~Den{\oe}ux, O.~Kanjanatarakul, and S.~Sriboonchitta, ``{EK-NN}clus: A
  clustering procedure based on the evidential $k$-nearest neighbor rule,''
  \emph{Knowledge-Based Systems}, vol.~88, pp. 57--69, 2015.

\bibitem{liu2015label}
K.~Liu, J.~Huang, H.~Sun, M.~Wan, Y.~Qi, and H.~Li, ``Label propagation based
  evolutionary clustering for detecting overlapping and non-overlapping
  communities in dynamic networks,'' \emph{Knowledge-Based Systems}, vol.~89,
  pp. 487--496, 2015.

\bibitem{danon2005comparing}
L.~Danon, A.~Diaz-Guilera, J.~Duch, and A.~Arenas, ``Comparing community
  structure identification,'' \emph{Journal of Statistical Mechanics: Theory
  and Experiment}, vol. 2005, no.~09, p. P09008, 2005.

\bibitem{ucidatasets}
``The {UCI} network data repository,''
  \url{http://networkdata.ics.uci.edu/index.php}.

\end{thebibliography}
\end{document}